\documentclass[letterpaper]{article} %
\usepackage{aaai25}  %
\usepackage{times}  %
\usepackage{helvet}  %
\usepackage{courier}  %
\usepackage[hyphens]{url}  %
\usepackage{graphicx} %
\usepackage{natbib}  %
\usepackage{caption} %
\usepackage{algorithm}
\usepackage{algpseudocode}
\usepackage{amsmath}
\usepackage{enumitem}
\usepackage{booktabs}
\usepackage{csquotes}
\usepackage{amssymb}
\usepackage{newfloat}
\usepackage{listings}
\DeclareCaptionStyle{ruled}{labelfont=normalfont,labelsep=colon,strut=off} %
\floatstyle{ruled}
\newfloat{listing}{tb}{lst}{}
\floatname{listing}{Listing}
\title{When Should We Prefer State-to-Visual DAgger Over Visual Reinforcement Learning?}

\author {
    Tongzhou Mu\textsuperscript{\rm 1}\equalcontrib, Zhaoyang Li\textsuperscript{\rm 1}\equalcontrib, Stanisław Wiktor Strzelecki\textsuperscript{\rm 1}\equalcontrib,\\
    Xiu Yuan\textsuperscript{\rm 1}, Yunchao Yao\textsuperscript{\rm 1}, Litian Liang\textsuperscript{\rm 1}, Hao Su\textsuperscript{\rm 1}
}
\affiliations {
    \textsuperscript{\rm 1} University of California San Diego\\
    \{t3mu, zhl165, sstrzelecki, x1yuan, y8yao, l6liang, haosu\}@ucsd.edu
}

\usepackage{bibentry}

\begin{document}

\newcommand{\method}{State-to-Visual DAgger}

\maketitle

\begin{abstract}

Learning policies from high-dimensional visual inputs, such as pixels and point clouds, is crucial in various applications. 
Visual reinforcement learning is a promising approach that directly trains policies from visual observations, although it faces challenges in sample efficiency and computational costs. 
This study conducts an empirical comparison of State-to-Visual DAgger — a two-stage framework that initially trains a state policy before adopting online imitation to learn a visual policy — and Visual RL across a diverse set of tasks. 
We evaluate both methods across 16 tasks from three benchmarks, focusing on their asymptotic performance, sample efficiency, and computational costs.
Surprisingly, our findings reveal that State-to-Visual DAgger does not universally outperform Visual RL but shows significant advantages in challenging tasks, offering more consistent performance. In contrast, its benefits in sample efficiency are less pronounced, although it often reduces the overall wall-clock time required for training.  
Based on our findings, we provide recommendations for practitioners and hope that our results contribute valuable perspectives for future research in visual policy learning.

\end{abstract}

\begin{links}
\link{Code}{https://github.com/tongzhoumu/s2v-dagger}
\end{links}

\section{Introduction}
\label{sec:intro}

Learning policies from high-dimensional visual observations, such as pixels and point clouds, is a crucial problem in fields like robotic manipulation \citep{nair2018visual, ze2024h, hansen2022temporal}, navigation \citep{gu2022multi}, and autonomous driving \citep{hossain2023autonomous}. 
Visual reinforcement learning (RL) methods, which employ RL algorithms on visual observations, stand out as a leading approach for acquiring such visual policies. 
Despite their popularity, visual RL methods are generally more prone to issues related to sample efficiency and computational costs than their counterparts utilizing low-dimensional state observations  \citep{chen2023visual}. 
This is primarily because visual RL must address two challenges \textit{concurrently}: 1) figuring out how to solve the task through trial and error; and 2) building a mapping from high-dimensional visual observations to the high-rewarding actions, a process that often involves training a large visual encoder.

\begin{figure*}[t]
    \centering
    \includegraphics[width=\textwidth]{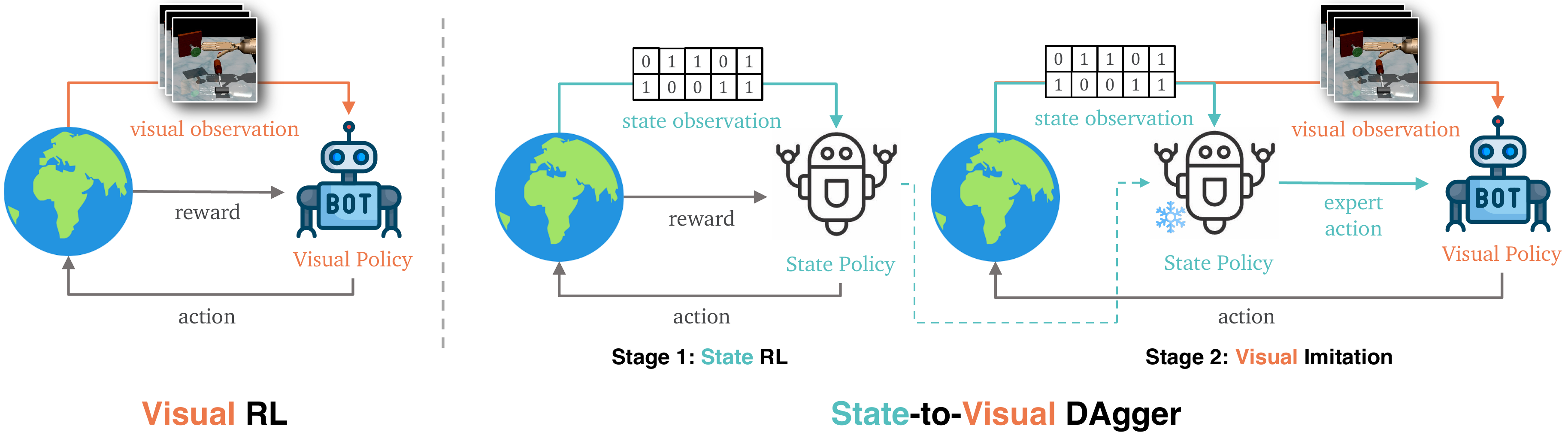}
    \caption{\textbf{Overview of Methods.} 
    While Visual RL directly trains a visual policy using RL, \method\ adopts a two-stage process: initially training a teacher policy with low-dimensional state observations, followed by teaching a visual policy via online imitation learning.}
    \label{fig:s2v}
\end{figure*}

A potential simplification of this problem is to tackle the two aforementioned challenges separately. Previous studies have utilized a two-stage approach for learning a visual policy, as illustrated in Fig. \ref{fig:s2v}. 
In the first stage, a teacher policy is trained using RL with low-dimensional state observations, possibly incorporating privileged information to facilitate learning. 
In the second stage, a visual policy is learned by online imitating the teacher policy, akin to DAgger \citep{ross2011reduction}. 
This two-stage framework has been applied across various applications, including dexterous manipulation \citep{chen2022system, chen2023visual}, legged locomotion \citep{lee2020learning, miki2022learning, zhuang2023robot, margolis2021learning}, drone control \citep{loquercio2021learning}, and autonomous driving \citep{chen2020learning}. 
In our study, this two-stage framework is referred to as "State-to-Visual DAgger", highlighting the transition from \textit{low-dimensional state observations} to \textit{high-dimensional visual observations}.

While \method\ can simplify the learning of visual policies by isolating focus at each stage, the added stage complicates training and may increase costs compared to single-stage visual RL methods. Therefore, our study explores the question: \textit{\textbf{When should \method\ be preferred over visual RL?}}

We investigate this question empirically by comparing \method\ against visual RL across diverse tasks and evaluation metrics. We selected \textbf{16} tasks from three benchmarks: ManiSkill \cite{gu2023maniskill2}, DMControl \cite{tassa2018diego}, and Adroit \cite{rajeswaran2017learning}. These tasks include stationary robot arm manipulation, mobile manipulation, dual-arm coordination, locomotion across different robot morphologies, classical control, and dexterous hand manipulation. Our comparison evaluates \textit{asymptotic performance, sample efficiency, and computational cost}, offering a comprehensive assessment of both methods.

The fairness of the comparison also hinges on the implementations. Despite its usage in several publications, a standardized implementation of \method\ has yet to be established. We meticulously developed our implementation of it, pinpointing several critical design decisions that significantly influence its performance. Further details on this are elaborated in Sec. \ref{sec:s2v_method}. Our evaluation revealed that \method\ does not consistently outperform visual RL, with key findings summarized below.

\textbf{Regarding asymptotic performance:} 
State-to-Visual DAgger demonstrates significant advantages over visual RL in hard tasks, but only achieves similar or slightly worse performance in easy tasks.
Notably, State-to-Visual DAgger usually provides more consistent and stable performance upon convergence.

\textbf{Regarding efficiency:} 
In scenarios where both \method\ and visual RL are capable of effectively solving tasks, \method\ does not distinctly outperform visual RL in terms of sample efficiency. Nevertheless, \method\ significantly improves wall-clock efficiency across the most tasks.

For a more in-depth discussion of our findings, please refer to Sec. \ref{sec:results}. Based on these empirical results, we also offer guidance for practitioners in Sec. \ref{sec:discussion}. \textbf{Our contributions} can be summarized as follows:

\begin{itemize}[leftmargin=*,nosep]
    \item We delve into the crucial question of "when \method should be preferred over visual RL," facilitated by a detailed comparison of a diverse set of tasks.
    \item Our analysis offers key insights and practical guidance for researchers and practitioners in visual policy learning.
    \item We offer a standardized implementation of \method\ and meticulously analyze several key design choices that significantly influence its performance.
\end{itemize}

\section{Related Works}

\subsubsection{Visual Reinforcement Learning:} Visual reinforcement learning (visual RL) integrates complex visual inputs, such as pixels and point clouds, into reinforcement learning algorithms, enabling agents to make decisions based on these observations. Visual RL can be categorized into model-free and model-based approaches. Model-free methods are divided into value-based and policy-based approaches. Value-based methods, such as those in \citep{mnih2015human, silver2013playing}, combine Q-learning with deep neural networks to learn from raw pixel inputs using convolutional neural networks. Policy-based methods, including \citep{klimov2017john, haarnoja2018soft}, optimize agents using policy gradients. For model-based visual RL, the agent needs to learn a world model from visual observations. Approaches such as PlaNet \citep{hafner2019learning}, Dreamer \citep{hafner2019dream}, Dreamer-v2 \citep{hafner2020mastering}, and TD-MPC \citep{hansen2022temporal} focus on learning dynamics from images and planning actions in latent spaces, with enhancements for discrete and continuous environments. Representation learning enhances visual RL performance, with prior works exploring pre-training using single-view \citep{shah2021rrl, parisi2022unsurprising}, multi-view \citep{driess2022reinforcement}, and video data \citep{kulkarni2019unsupervised}. Additionally, self-supervised learning \citep{laskin2020curl} and data augmentation \citep{yarats2020image} enhance performance without pre-training. Practical applications include QT-Opt \citep{kalashnikov2018scalable} for real-world robotics manipulation and Akkaya et al.'s work enabling a robotic hand to solve a Rubik's Cube \citep{akkaya2019solving}. However, visual RL faces challenges in sample efficiency and computational costs compared to low-dimensional approaches \citep{chen2023visual}, and it struggles with computational efficiency and generalizability across different visual domains.

\subsubsection{Utilize Privileged Information During RL Training:}
Privileged information can accelerate visual RL learning and improve sampling efficiency. While unavailable during deployment, it is often accessible during training and can be strategically utilized. For example, \citet{kaufmann2023champion} uses privileged information about the highly accurate simulation of drone dynamics and environment and optimal race routes to help RL models train more effectively. Some methods, such as those described in \citep{pinto2017asymmetric, kumar2107rma}, utilize simulation information to provide detailed and controlled feedback on actions within a simulated environment, thus enhancing the robustness of the RL policy. Besides using available privileged information during RL, some methods follow the teacher-student approach we call \method, such as training the policy as the expert and then using the privileged information from the expert to supervise the student model. 
\method\ has been used in applications about autonomous driving \citep{chen2020learning}, legged locomotion \citep{lee2020learning, miki2022learning, zhuang2023robot}, drone control \citep{loquercio2021learning}, dexterous grasp \citep{xu2023unidexgrasp}, and dexterous manipulation \citep{chen2023visual}, which utilize the privileged information to depth. Although previous work does not investigate whether this \method improves learning efficiency, we focus on investigating the learning efficiency of \method\, compared to single-stage visual RL, and clarify what situation we need to use \method.

\section{Methods}

Our study aims to conduct a comparison between two distinct paradigms to learning visual policies: State-to-Visual DAgger and visual Reinforcement Learning. To provide a focused examination, we focus on representative methods within each paradigm.
Given the absence of a standardized open-source implementation of State-to-Visual DAgger, we have carefully developed our version, identifying several crucial design choices that significantly affect its performance.
Visual RL encompasses a wide range of approaches, each with its own strengths. For a fair comparison, we chose Asymmetric Actor Critic \citep{pinto2017asymmetric} as the visual RL counterpart in this study. This method was selected due to its ability to incorporate privileged state information, similar to the advantage used by State-to-Visual DAgger. This section details the design and implementation of these methods.

\subsection{State-to-Visual DAgger}
\label{sec:s2v_method}
State-to-Visual DAgger adopts a two-stage approach to learning visual policies, as depicted in Fig. \ref{fig:s2v}. This method requires the training environment to \textit{concurrently} supports two observation spaces: a low-dimensional state observation space denoted as $\mathbb{O}^\mathcal{S}$, and a high-dimensional visual observation space $\mathbb{O}^\mathcal{V}$. This approach usually assumes training in a simulator, which offers both the full system state and rendered images. However, the final visual policy learned by \method does not rely on any simulator-specific privileged information.

\subsubsection{Stage 1: Learning State Policy by RL.}
In the initial stage, we employ Soft Actor-Critic (SAC) \citep{haarnoja2018soft}, a widely used RL algorithm, to train state-based a teacher policy $\pi^{\mathcal{S}}$ using low-dimensional state observations. 
Here, state observation refers to a low-dimensional vector that describes the current state, often incorporating privileged information not available during real-world policy deployment.
For instance, in the PickCube task from ManiSkill, the state observation includes both the robot's proprioception data and the ground truth pose of the cube. While the robot proprioception data can be accessible in the real world, the ground truth pose of the cube typically is not. 
Our experiments directly employ the low-dimensional state observations provided by the environment's interface (more details in the Appendix \ref{sec:appendix-tasks}). 
The learned state-based teacher policy will guide the subsequent learning process of visual policy.
In our experiments, the training of stage 1 is stopped upon convergence, and we save the latest checkpoint. Alternatively, the final checkpoint could be selected based on a predetermined computational budget.

\subsubsection{Stage 2: Learning Visual Policy by DAgger.}
In Stage 2, the state policy acquired from Stage 1 serves as a teacher to guide the learning of the visual policy. This is achieved by using DAgger \citep{ross2011reduction}, an online imitation learning algorithm. DAgger's primary advantage over traditional offline imitation methods lies in its ability to mitigate the covariate shift problem by leveraging an expanding from online interactions. The training of the visual policy $\pi^{\mathcal{V}}$ is done by minimizing the MSE loss on actions, formulated as:
\begin{equation}
    \pi^{\mathcal{V}}=\text{argmin}_{\pi^{\mathcal{V}}} \lVert \pi^{\mathcal{V}}(o_t^{\mathcal{V}})-\pi^{\mathcal{S}}(o_t^{\mathcal{S}})\rVert^2 
    \label{eq:bc_loss}, 
\end{equation}
where $o_t^{\mathcal{V}}$ and $o_t^{\mathcal{S}}$ are paired visual observation and state observation. Our implementation of DAgger for learning visual policies incorporates two critical design decisions:
\begin{enumerate}
    \item DAgger can be implemented in both on-policy and off-policy manners, analogous to the methods used in on-policy and off-policy reinforcement learning. The primary distinction lies in whether to incorporate off-policy trajectories into the training dataset via a replay buffer. Our experiments demonstrate that the off-policy version significantly outperforms the on-policy variant, likely due to its ability to retain a more diverse set of training examples.
    
    \item Rather than employing a fixed number of gradient updates per training round, we utilize an early-stopping mechanism triggered when a predefined imitation loss threshold is reached. After early stopping, a new cycle of data collection is initiated through interaction with the environment, followed by the integration of this new data into the buffer. This approach reduces unnecessary training on patterns that have already been learned, thereby preventing overfitting and enhancing training efficiency.
\end{enumerate}

For a detailed description of our State-to-Visual DAgger implementation, please see Algorithm \ref{alg:s2v}. Further implementation details can be found in Appendix \ref{sec:appendix-implementation-details}.

\begin{figure*}[t]
    \centering
    \includegraphics[width=\textwidth]{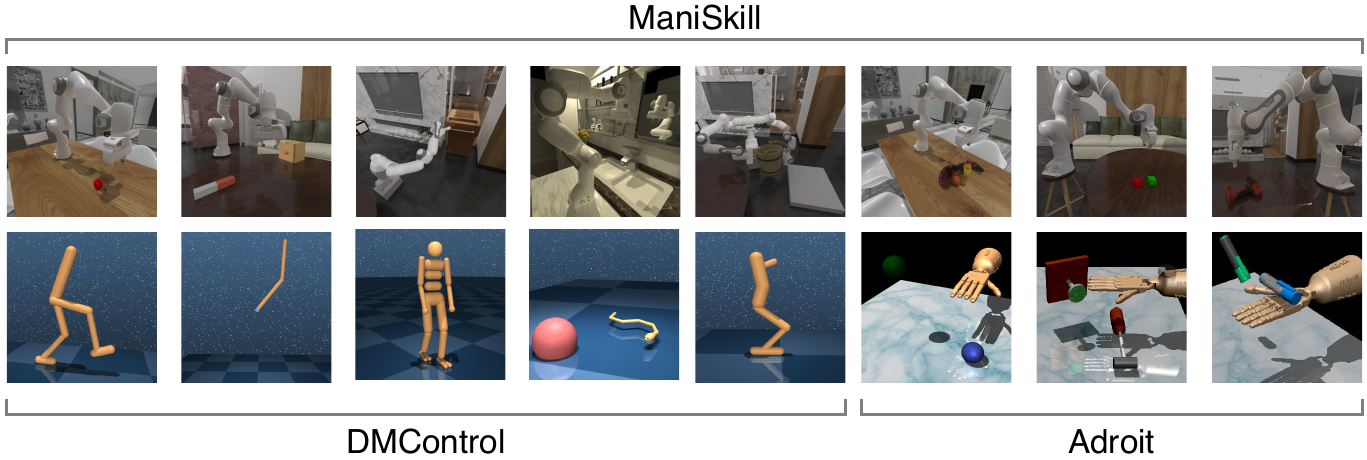}
    \caption{\textbf{Examples of Tasks.} 
    We consider control tasks spanning $\mathbf{3}$ benchmarks. The first row contains tasks from ManiSkill (stationary and mobile robot arm manipulation, dual-arm coordination). The first five tasks in the second row are from DMControl (various robot morphologies for locomotion and classical control tasks), and the remaining three tasks in the second row are from Adroit (dexterous hand manipulation).}
    \label{fig:tasks}
\end{figure*}

\subsection{Visual Reinforcement Learning}

While numerous visual RL algorithms exist \citep{impala,drq,drq2,planet}, a direct comparison with the State-to-Visual DAgger method may not be entirely fair. This discrepancy arises because standard visual RL approaches do not leverage the privileged information that State-to-Visual DAgger capitalizes on, a factor that substantially aids in solving the tasks.

To ensure a more balanced comparison, we adopt the Asymmetric Actor Critic \citep{pinto2017asymmetric} as the visual RL method for our study. This algorithm uniquely lets the critic take the state (including privileged information) as input, whereas the actor still operates on high-dimensional visual inputs. This design enables the utilization of privileged information without making the policy dependent on it.
Originally, the Asymmetric Actor Critic employed DDPG \citep{ddpg} as its underlying RL algorithm; however, we opted for SAC to enhance performance. We refrained from incorporating advanced techniques for image-based feature extraction \citep{flare} and data augmentation \citep{drq2} that have been recently introduced, as their application to both State-to-Visual DAgger and visual RL would unlikely change the core findings of our study significantly.

Our empirical evaluations show that Asymmetric Actor Critic, when combined with SAC, matches the performance of state-of-the-art visual RL algorithms on the tasks we tested.This justifies its selection as the representative visual RL method for our comparisons. Details are in Appendix \ref{sec:appendix_drq}.

\section{Experimental Setup}
\label{sec:experimental_setup}
Our experimental setup is designed to thoroughly evaluate and compare the capabilities of two methods for learning visual policies, spanning a diverse range of tasks and employing specific evaluation metrics to gauge performance comprehensively. We discuss the details in this section.

\subsection{Task Descriptions}
Our experiments span \textbf{16} tasks across \textbf{3} benchmarks: ManiSkill (robotic manipulation; 8 tasks), DMControl (locomotion and control; 5 tasks), and Adroit (dexterous manipulation; 3 tasks). This diverse set includes stationary and mobile robot arm manipulation, dual-arm coordination, various robot morphologies for locomotion, classical control, and dexterous hand manipulation. The range ensures our conclusions are comprehensive and unbiased. Figure \ref{fig:tasks} illustrates all \textbf{16} tasks. Detailed task descriptions and setups are provided in Appendix \ref{sec:appendix-tasks}, summarized as follows:

\subsubsection{ManiSkill:} Features robotic manipulation tasks where low-dimensional state observations include robot proprioception (joint angles, joint velocities, end effector pose, base pose, etc.) and ground truth object or goal information, with visual observations from dual 64$\times$64 RGBD cameras.

\subsubsection{DMControl:} We evaluate on locomotion and classical control tasks, following standard protocols \citep{drq, drq2, planet}. State observations primarily include robot proprioceptive data. Visual inputs are 84$\times$84 RGB images, stacking 3 frames. We adopt action repeat parameters from \citep{drq}.

\subsubsection{Adroit:} Concentrates on dexterous manipulation tasks, with low-dimensional state observations detailing the information about all the joints as well as the pose of the palm and poses of other objects in the environment. Visual observations are 128$\times$128 RGB images.

\subsection{Evaluation Metrics}
\label{sec-subsection-evaluation-metrics}
Our comparison of visual policy learning methods centers on two evaluation metrics: learning efficiency and asymptotic performance.

\begin{figure*}[h]
    \centering
    \includegraphics[width=\textwidth]{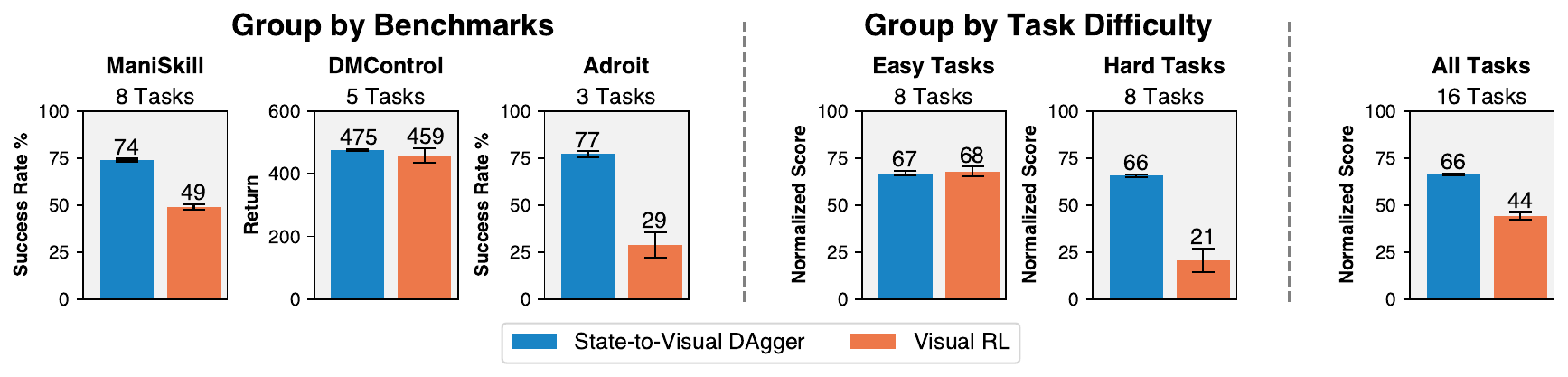}
    \caption{
    \textbf{Performance Overview.} The figure features histograms comparing average performance across different dimensions. On the left, three histograms present performance by benchmark (success rates for ManiSkill and Adroit, and returns for DMControl). In the center, two histograms categorize performance by task difficulty, utilizing normalized scores (success rate for ManiSkill and Adroit, return divided by 1000 for DMControl) to accommodate the varying metrics across benchmarks.
    The error bars represent the 95\% CI over three seeds.
    }
    \label{fig:overall_performance}
\end{figure*}

\subsubsection{Learning Efficiency:} We evaluate efficiency in terms of both sample efficiency (gauged by the number of environment steps) and computational cost (measured in wall-clock time), considering the cumulative costs of the two stages in \method for a balanced comparison. All experiments are standardized on the same hardware to ensure fair comparisons of computational costs. Our hardware setting: 32 CPU cores (Intel Xeon 2.1GHz) and 1 GPU (NVIDIA-GeForce-RTX-2080-Ti with 11GB).

\subsubsection{Asymptotic Performance:} To address the challenge of calculating asymptotic performance in RL experiments, we average data points over a window at the end of the learning curve to gauge this metric, with the window set at 3\% of total environment steps.

\begin{figure*}[t]
    \centering
    \vspace{-0.3 cm}
    \includegraphics[width=\textwidth]{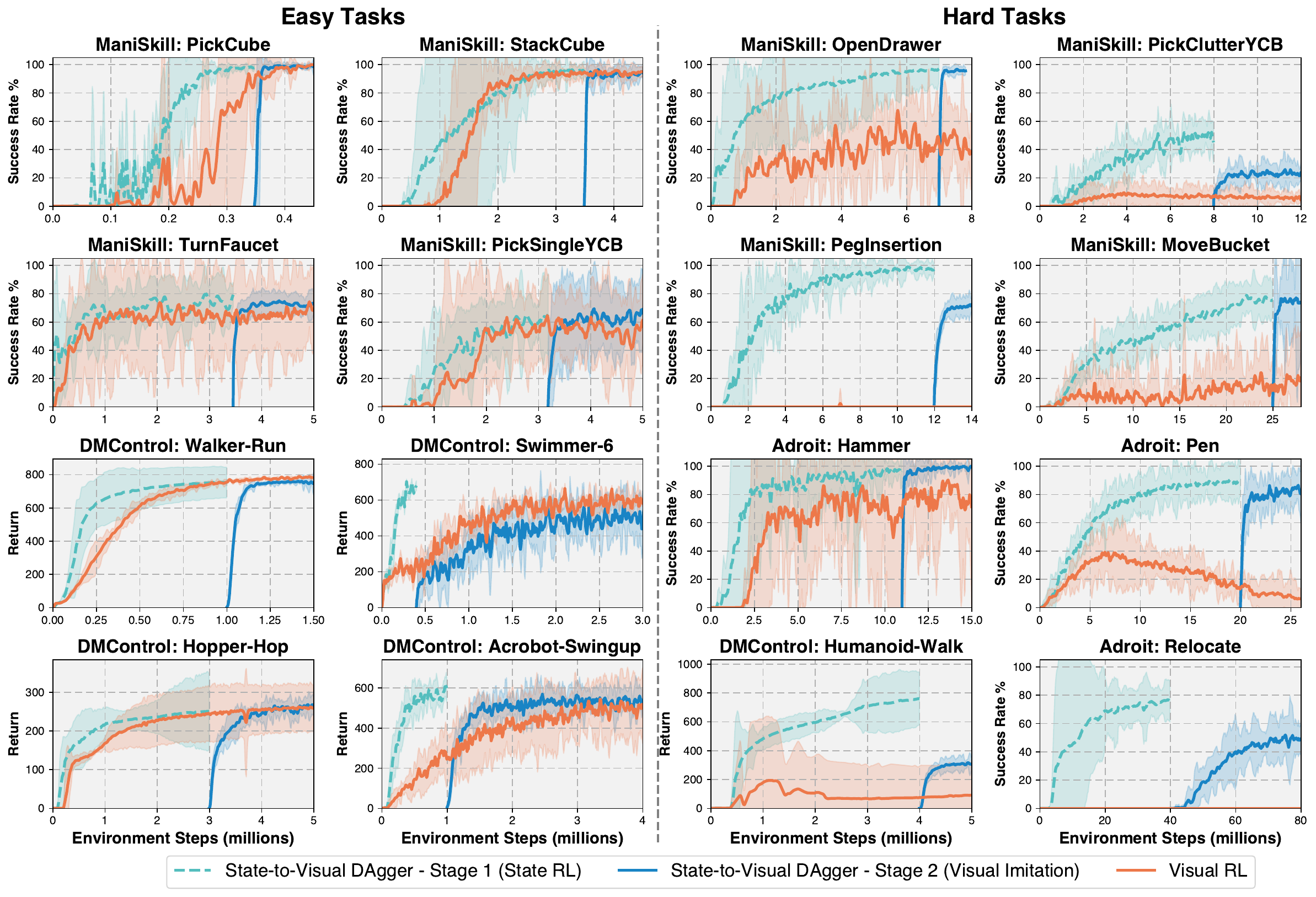}
    \caption{
    \textbf{Learning curves against environment steps.} 
    Success rate (ManiSkill and Adroit) and return (DMControl) in each task. 
    Tasks are categorized as easy if state-based RL converges within 4M steps, while the others are considered hard. 
    State-to-Visual DAgger (Stage 2) comparisons with visual RL should account for the cost of Stage 1. 
    The curve for stage 1 serves as a reference but is not directly comparable to others due to its state-based policy nature. 
    The shaded region represents the 95\% CI across three seeds.
    } 
    \label{fig:environment_steps}
\end{figure*}

\section{Results}
\label{sec:results}
In this section, We analyze the experimental results, highlighting key findings, with detailed implications and advice for practitioners in Sec. \ref{sec:discussion}. All experiments use three random seeds, aggregating results across tasks for reliability.

\subsection{Performance Comparison}
\label{sec:performance}
Our results suggest that the efficacy of \method\ compared to visual RL varies across tasks, as illustrated in Fig. \ref{fig:overall_performance} and Fig. \ref{fig:environment_steps}. \textit{There is no single approach that consistently outperforms the other across all tasks.} Specifically, \method\ shows notable superiority in many tasks within the ManiSkill and Adroit benchmarks. Conversely, visual RL exhibits marginal benefits in the majority of tasks from the DMControl benchmark.

Given that previous works mainly apply \method\ to exceptionally challenging tasks, such as dexterous manipulation ~\cite{chen2023visual} and drone control ~\cite{loquercio2021learning}, categorizing tasks by their difficulty level may offer a clearer perspective. 
Here, we define "easy tasks" as those where state-based RL achieves convergence within 4 million environment steps, with all other tasks classified as "hard". Although this classification is not rigorous, it facilitates a more detailed comparison between \method\ and visual RL. 
As illustrated in Fig. \ref{fig:overall_performance}, \textit{\method\ markedly surpasses visual RL in hard tasks, but only achieves similar or slightly worse results in easier tasks.} 
The learning curves for each task, shown in Fig. \ref{fig:environment_steps}, further validate this observation. While \method\ excels at difficult tasks through imitation of state policies, its performance on easier tasks is comparable to or slightly below that of visual RL.

The performance gap in \textit{hard tasks} stems from the disparity between state-based and visual RL. State-based RL with a simple MLP handles these tasks well (with dense rewards), while visual RL struggles. We hypothesize that noisy gradients during exploration impede CNN training with image observations.

\subsection{Consistency and Stability}
Our results also indicate that \textit{\method\ delivers more consistent performance at convergence}, as evidenced by the narrower confidence intervals observed across all benchmarks and difficulty levels, as illustrated in Fig. \ref{fig:overall_performance}.  

A closer look at individual task performances, as shown in Fig. \ref{fig:environment_steps}, further shows that visual RL may exhibit fluctuating performance on certain tasks (e.g., ManiSkill OpenDrawer and Adroit Hammer), and may even unlearn (e.g., Adroit Pen). 
Conversely, \textit{the performance of \method\ (Stage 2) remains more stable upon convergence}, as indicated by the smoother learning curves. This stability is expected, as imitation learning in Stage 2 is inherently easier and more stable than reinforcement learning.It simplifies learning and streamlines deployment checkpoint selection.

\begin{figure*}[t]
    \centering
    \includegraphics[width=\textwidth]{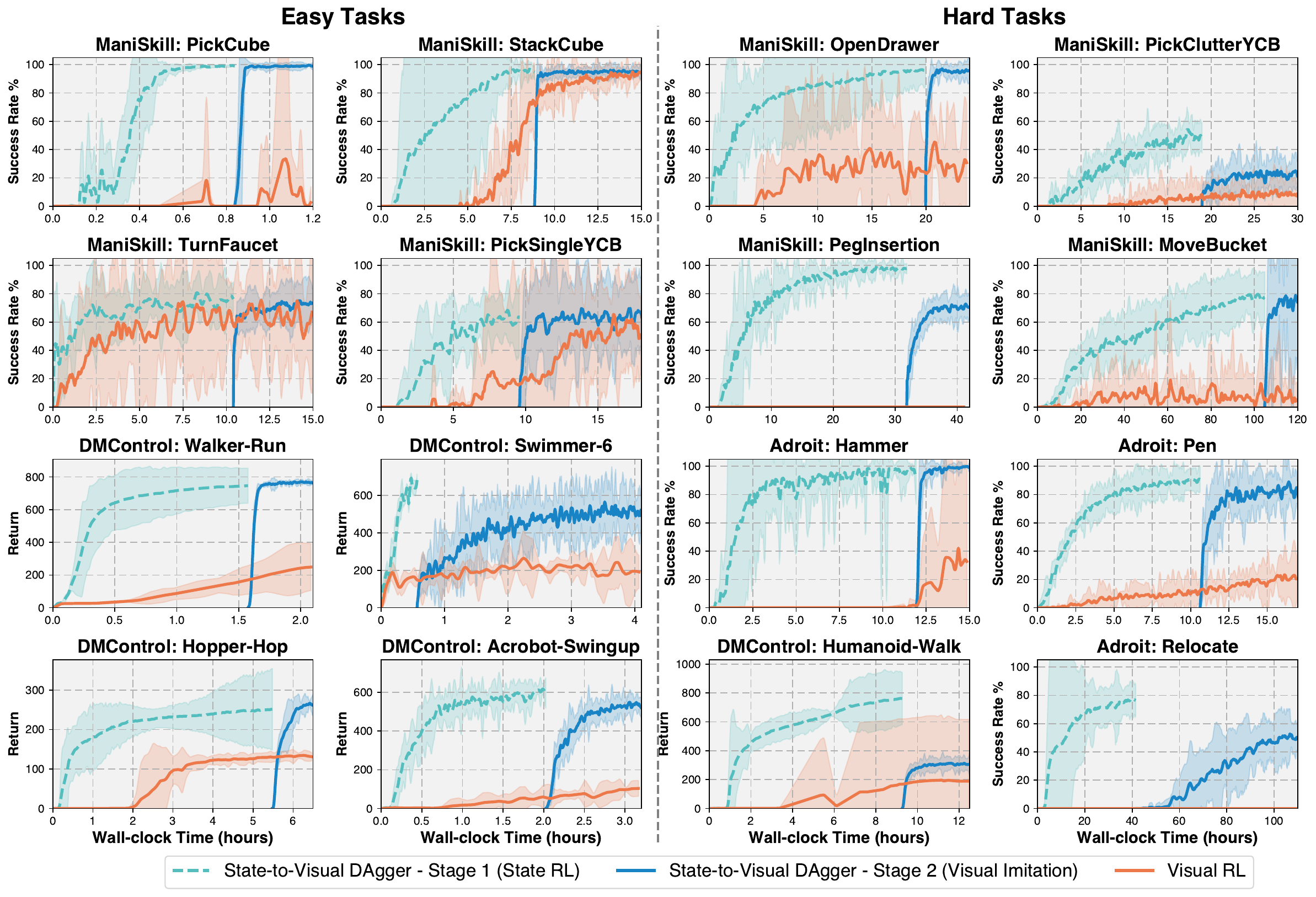}
    \caption{\textbf{Wall-clock Time.} Similar to Fig. \ref{fig:environment_steps}, however, we use the wall-clock time as the x-axis instead of the environment steps. 
    We find that \method\ has better wall-clock time efficiency than visual RL on most tasks. 
    }
    \label{fig:wall_time}
\end{figure*}

\subsection{Sample efficiency (Environment Steps)}
\label{sec:sample_efficiency}

Comparing the sample efficiency of \method\ and visual RL is not straightforward due to the inherent structure of \method, where a visual policy is not trained until the second stage. 
Observations in Fig. \ref{fig:environment_steps} suggests \method\ appears more sample-efficient than visual RL in hard tasks, primarily due to its superior asymptotic performance rather than true sample efficiency.

Conversely, when it comes to easier tasks where both methods converge to similar levels of performance, \method\ does not demonstrate a clear advantage in sample efficiency over visual RL. This leads to the conclusion that the apparent higher sample efficiency of \method\ in certain scenarios is more attributed to its enhanced asymptotic performance rather than an intrinsic efficiency advantage. Thus, \textit{when both methods are capable of effectively solving tasks, \method\ does not offer significant benefits in sample efficiency over visual RL}.

\subsection{Computational Cost (Wall-clock Time)}
Although \method\ may not enhance sample efficiency, it excels in wall-clock time, consistently outperforming visual RL across all tasks as shown in Fig. \ref{fig:wall_time}.

We found that \textit{\method\ is notably more time-efficient than visual RL in the majority of tasks}, including those categorized as easy, where it did not show superior sample efficiency, as discussed earlier. The rationale for this efficiency is straightforward: visual RL involves the training of a visual encoder and the rendering of visual observations during training, both of which are time-consuming processes. \method\ requires these processes only in the second stage, while the first-stage state-based RL runs much faster in wall-clock time. This distinction underscores \method's significant time-saving benefits, even with comparable sample efficiency to visual RL. While factors like rendering speeds, network sizes, and hyperparameters affect wall-clock time and limit comparison rigor, the time-saving advantage of \method's first stage is expected to hold across settings.

\section{Discussions and Recommendations}
\label{sec:discussion}

Our analysis reveals that no single method uniformly surpasses the other in every task, highlighting the distinct strengths and limitations of each approach. Below, we provide guidance for practitioners in visual policy learning, derived from our empirical findings. It is important to note, however, that these recommendations are based on observations from our experiments and should be considered as informed suggestions rather than definitive rules.

\subsection{Recommend to Use \method}
\subsubsection{Visual RL Struggles to Solve the Task:} For challenging tasks where visual RL falls short, \method\ is preferred, leveraging low-dimensional state information for effective policy learning before transitioning to high-dimensional visual inputs.

\subsubsection{You Have Already Tried State RL:} If you have state RL implemented and can extract or simulate low-dimensional state observations, transitioning to \method\ is a natural next step, building on existing work without retraining a visual RL agent.

\subsubsection{Focus on Wall-Clock Time Efficiency:} For projects prioritizing computational cost and execution time, \method\ is the optimal choice. Our experiments show that \method\ significantly reduces wall-clock time compared to traditional visual RL methods, without compromising outcome quality.

\subsection{Recommend to Use Visual RL}

\subsubsection{Low-Dimensional State Observations Are Not Available:} If the environment does not provide, or it is not feasible to simulate, low-dimensional state observations necessary for the state-based teacher policy, visual RL becomes the more viable option. In such cases, direct learning from high-dimensional visual observations is the only path forward.

\subsubsection{Preference for Minimal Intervention During Training:} Visual RL provides a straightforward, hands-off approach to policy training, avoiding intermediate steps like interrupting state RL training to select checkpoints and switching to visual imitation. For a process requiring less intervention and manual oversight, visual RL may better suit your workflow.

\subsubsection{Tasks Evidently Solvable by Visual RL:} For simpler tasks where visual RL has been shown to be effective, starting with visual RL might be the most practical choice. It simplifies the setup process by removing the need for a two-stage training protocol and can achieve performance on par with \method\ in these scenarios, making it an efficient and straightforward solution.

\section{Conclusions}
\label{sec:conclusions}
Our research compares \method\ and visual RL on asymptotic performance, sample efficiency, and computational costs across tasks, highlighting their unique strengths and limitations to guide strategic application choices. We provide practical guidelines for selecting between \method\ and visual RL, considering task complexity and context to determine the preferred method. However, our study has several limitations. Firstly, the categorization of tasks as difficult based on a threshold number of environmental steps is not rigorous. Additionally, we did not investigate the impact of different checkpoint selections on \method's efficiency and performance, which could provide further insights into its adaptability. Future research should analyze checkpoint selection effects to ensure a fair and thorough comparison between \method\ and visual RL.

\bibliography{ref}
\newpage
\appendix
\section*{Appendix}
\section{Why \method\ is introduced in previous works?}
Visual reinforcement learning (visual RL) has two challenges. Firstly, visual RL has to learn which feature to extract from visual observations. Secondly, visual RL needs to learn what high-rewarding actions are ~\cite{chen2023visual}. To summarize, two difficulties exist: learning how to observe and act ~\cite{andrychowicz2020learning}. \method\ intuitively splits two tasks into two stages to be accomplished. The first stage handles learning how to observe, and the second stage handles learning how to act. There are several advantages to using the \method. 

\begin{itemize}
    \item In the first stage,  teacher policy is trained by low-dimensional observations more efficiently using reinforcement learning (RL) ~\cite{zhou2019does, chen2023visual}.
    \item \method\ simplifies the difficulty of training a visual policy by learning of imitation of the teacher policy that has already been trained.
    \item \method\ eases the high-level controller from being affected by complex joint-level drives, while the low-level controller does not need to infer from visual observations ~\cite{margolis2021learning}.
    \item It facilitates distributed learning, more specifically, since the trained state policy in the first stage is a "white box," which reveals all internal states so it can be simultaneously in any environment state for every possible instruction in the second stage.
\end{itemize}

\section{Task Descriptions}
\label{sec:appendix-tasks}

\subsection{ManiSkill}

\label{sec:appendix-tasks-ManiSkill}
In this section, we explore \textbf{8} tasks derived from the ManiSkill2 benchmark, namely PickCube, StackCube, PickSingleYCB, PickClutterYCB, PegInsertionSide, TurnFaucet, OpenDrawer, and MoveBucket. These tasks are designed to emulate manipulation challenges of varying degrees of difficulty and are characterized by their meticulously engineered dense rewards. Each task utilizes 7 DoF Panda robotic arms, with the setup predominantly featuring a stationary single arm. Exceptions include OpenCabinetDrawer, which involves a mobile robot equipped with one such arm, and MoveBucket, where the robot is equipped with two arms. For state observation, we have the full state of the robot (joint angles, pose of the end effector, pose of the base, etc.), and the poses of other objects and goals, depending on the task. For visual observations, we have  $64 \times 64$ RGBD images rendered from two different cameras. For OpenCabinetDrawer and MoveBucket, we have a $125 \times 50$ RGBD image rendered from a panoramic camera mounted on the robot. We describe each task as follows:
\begin{itemize}
    \item 
        PickCube: The task is defined as a basic manipulation challenge where the objective is to grasp a cube located at a random position and elevate it to a specified target location. Success is achieved when the cube is positioned within a 2.5 cm radius of the target location and the robot remains stationary.
    \item 
        StackCube: The manipulation task is designed with the goal of picking up a red cube located at a random position and placing it atop a green cube. Success criteria are met when the red cube is stably positioned on the green cube without being held, indicating effective manipulation and placement skills.
    \item 
        PickSingleYCB: A manipulation task with the objective of picking up a randomly selected object. The object is one of the YCB benchmark objects ~\cite{calli2015ycb}, simulating real-life objects. We succeed if the object is within 2.5cm of the goal position and the robot is static.
    \item 
        PickClutterYCB: A manipulation task with the objective of picking up a randomly selected YCB object. This time there are 4-8 objects lying down and we have to pick up the right one. 
    \item 
        TurnFaucet: A manipulation task with the objective of turning one of 60 predefined faucets. The robot should grab the handle of the faucet and turn it past a target angular distance.
    \item 
        PegInsertionSide: A manipulation task with the objective of picking up a cuboid-shaped peg and then placing it into a hole in a box. We succeed if half of the peg is inside the hole. One of the difficulty factors is that it requires a high precision to fit the peg, as the hole has a small margin towards the peg size.
    \item 
        OpenDrawer: A manipulation task with the objective of opening a cabinet drawer. There are multiple drawer models, the used one being arbitrarily chosen from them. The robot should successfully open the door of the drawer attached by a prismatic joint. We succeed if the drawer is open to at least 90\% of its range and is static.
    \item 
        MoveBucket: A manipulation task with the objective of lifting a bucket with a ball inside and placing it on a platform. There are 29 models of buckets use for training. 
        We succeed if we place the bucket on the platform in an upright position, it is static and the ball remains inside. This task is very difficult, as it requires two-arm coordination and the ball inside makes the center of mass constantly change.
\end{itemize}
We refer to ~\cite{mu2021maniskill, gu2023maniskill2} for additional details.

\subsection{DMControl}
\label{sec:appendix-tasks-DMControl}
We consider \textbf{5} tasks from the DMControl suite: Acrobot-Swingup, Walker-Run, Hopper-Hop, Swimmer-6, and Humanoid-Walk, which represent continuous control tasks of varying difficulty. These tasks vary in embodiment, objective,
action space, and reward type. While the states (and so the state observations) vary between environments, all visual observations are pixels of a $84\times 84$ image. 
We describe each task as follows:
\begin{itemize}
    \item
        Acrobot Swingup: A control problem with a planar, underactuated (1 DoF) double pendulum. The goal is to swing up and balance. There is a smooth reward depending on the pendulum's position.
    \item 
        Hopper Hop: A locomotion task with a planar one-legged 4 DoF Hopper. The goal is to hop forward, reaching a decent velocity. The reward is dependent on the forward velocity and the torso height.
    \item
        Walker Run: A locomotion task with a planar Walker embodiment, having 6 DoF. The
        task is to run forward at a high velocity until the end of the episode. There is a dense (shaped) reward that varies with forward velocity and positioning of joints.
    \item
        Swimmer-6: A locomotion task with a 6-link planar Swimmer, having 5 DoF. The goal is to have the swimmer's nose inside a target sphere. The reward for achieving the target is +1 and smoothly decreases with the distance from the sphere.
    \item 
        Humanoid Walk: The most complex locomotion task out of these six, consists of a simplified 3D humanoid with 21 joints. The task is to walk, so achieve a horizontal speed of 1m/s. 
\end{itemize}
We refer to ~\cite{tassa2018diego} for additional task details.

\subsection{Adroit}
\label{sec:appendix-tasks-Adroit}
The Adroit benchmark consists of problems that should be solved using a complex, 24-DoF manipulator, simulating a real hand. We consider \textbf{3} tasks from that suit: Relocate, Pen, and Hammer. Each of these tasks has additional degrees of freedom for the movement of the arm (4 extra for most of the tasks, 6 extra for Relocate). The state observation consists of information about all the joints as well as the pose of the palm and poses of other objects in the environment. The visual observation is $128 \times 128$ RGB images. All environments use a dense reward depending on our progress towards the objective. 
We describe each task as follows:
\begin{itemize}
    \item Relocation: We have to pick up and move a ball to a random target location. 
    \item Pen: We have a pen in our hand, and we have to reposition it so it matches a given (randomly chosen) orientation.
    \item Hammer: We are given a hammer tool, and we have to hammer a nail into a wall.
\end{itemize}
We refer to ~\cite{rajeswaran2017learning} for additional task details. 

\section{Implementation Details}
\label{sec:appendix-implementation-details}

\subsection{\method}

\method uses different Convolutional Neural Networks (CNNs) for other tasks with benchmarks. We show the details of CNNs separating into \textbf{3} parts for ManiSkill, DMControl, and Adroit.

For the tasks, OpenCabinetDrawer and MoveBucket in ManiSkill, the architecture of CNN is:

\vspace{1em}

{\begin{lstlisting}[basicstyle=\footnotesize\ttfamily, breaklines=true, frame=single, numbers=none]
(0): nn.Conv2d(in_channels, 16, 3, padding=1, bias=True), nn.ReLU(inplace=True),
(1): nn.MaxPool2d(2, 2),  # [25, 62]
(2): nn.Conv2d(16, 16, 3, padding=1, bias=True), nn.ReLU(inplace=True),
(3): nn.MaxPool2d(2, 2),  # [12, 31]
(4): nn.Conv2d(16, 32, 3, padding=1, bias=True), nn.ReLU(inplace=True),
(5): nn.MaxPool2d(2, 2),  # [6, 15]
(6): nn.Conv2d(32, 64, 3, padding=1, bias=True), nn.ReLU(inplace=True),
(7): nn.MaxPool2d(2, 2),  # [3, 7]
(8): nn.Conv2d(64, 128, 3, padding=1, bias=True), nn.ReLU(inplace=True),
(9): nn.Linear(128 * 3 * 7, out_dim), nn.ReLU(inplace=True)
\end{lstlisting}

\vspace{1em}

\begin{algorithm}[htbp]
    \caption{\textbf{State-to-Visual DAgger - Stage 2 (Visual Imitation)}}
    \label{alg:s2v}
    \small
    \begin{algorithmic}[1]
    \Require Task MDP $\mathcal{M}$, State policy $\pi^{\mathcal{S}}$ obtained in Stage 1, Number of samples to collect in each round $N_{collect}$, Loss threshold $\delta$ for early stopping
    \State Initialize Visual Policy $\pi^{\mathcal{V}}$, replay buffer $\mathcal{B}$
    \For {each round} 
        \State Collect $N_{collect}$ samples $\{(o_1^{\mathcal{S}}, o_1^{\mathcal{V}}), (o_2^{\mathcal{S}}, o_2^{\mathcal{V}}), ...\}$ by executing $\pi^{\mathcal{V}}$ in $\mathcal{M}$
        \Comment $o_t^{\mathcal{S}}$ and $o_t^{\mathcal{V}}$ are state and visual observations at step $t$, respectively 
        \State Compute expert actions: $\{a_1^{\mathcal{S}}, a_2^{\mathcal{S}}, ...~|~ 
a_t^{\mathcal{S}}=\pi^{\mathcal{S}}(o_t^{\mathcal{S}})\}$
        \Comment $a_t^{\mathcal{S}}$ will be the supervision in imitation
        \State Add visual observations and expert actions to buffer: $\mathcal{B} \leftarrow \mathcal{B} \cup \{(o_1^{\mathcal{V}}, a_1^{\mathcal{S}}), (o_2^{\mathcal{V}}, a_2^{\mathcal{S}}), ...\}$
        \For {each gradient step}
            \State Sample data from buffer $\mathcal{B}$
            \State Compute the imitation loss $J$ according to Eq. \ref{eq:bc_loss}
            \State Update visual policy $\pi^{\mathcal{V}}$ using the imitation loss $J$
            \If {$J<\delta$}
                \Comment early stop if the loss is smaller than the predefined threshold
                \State \textbf{break}
            \EndIf
        \EndFor

    \EndFor
    \end{algorithmic}
\end{algorithm}

For other tasks in ManiSkill, the architecture of CNN is as follows:

\vspace{1em}

\begin{lstlisting}[basicstyle=\footnotesize\ttfamily, breaklines=true, frame=single, numbers=none]
(0): nn.Conv2d(in_channels, 16, 3, padding=1, bias=True), nn.ReLU(inplace=True),
(1): nn.MaxPool2d(2, 2),  # [32, 32]
(2): nn.Conv2d(16, 16, 3, padding=1, bias=True), nn.ReLU(inplace=True),
(3): nn.MaxPool2d(2, 2),  # [16, 16]
(4): nn.Conv2d(16, 32, 3, padding=1, bias=True), nn.ReLU(inplace=True),
(5): nn.MaxPool2d(2, 2),  # [8, 8]
(6): nn.Conv2d(32, 64, 3, padding=1, bias=True), nn.ReLU(inplace=True),
(7): nn.MaxPool2d(2, 2),  # [4, 4]
(8): nn.Conv2d(64, 128, 3, padding=1, bias=True), nn.ReLU(inplace=True),
(9): nn.MaxPool2d(2, 2),  # [2, 2]
(10): nn.Conv2d(128, 128, 1, padding=0, bias=True), nn.ReLU(inplace=True),
(11): nn.Linear(128 * 4, out_dim), nn.ReLU()
\end{lstlisting}

\vspace{1em}

For tasks in DMControl, the architecture of CNN is:

\vspace{1em}

\begin{lstlisting}[basicstyle=\footnotesize\ttfamily, breaklines=true, frame=single, numbers=none]
(0): nn.Conv2d(in_channel, 32, 3, stride=2), nn.ReLU(),
(1): nn.Conv2d(32, 32, 3, stride=1), nn.ReLU(),
(2): nn.Conv2d(32, 32, 3, stride=1), nn.ReLU(),
(3): nn.Conv2d(32, 32, 3, stride=1), nn.ReLU(),
(4): flatten(1)
(5): nn.Linear(32 * 35 * 35, 256),
(6): nn.LayerNorm(256)
\end{lstlisting}

\vspace{1em}

For tasks in Adroit, the architecture of CNN that input channels equal to $3$ and output channels equal to $256$ is as follows:

\vspace{1em}

\begin{lstlisting}[basicstyle=\footnotesize\ttfamily, breaklines=true, frame=single, numbers=none]
(0): nn.Conv2d(in_channels, 16, 3, padding=1, bias=True), nn.ReLU(inplace=True),
(1): nn.MaxPool2d(2, 2),  # [64, 64]
(2): nn.Conv2d(16, 16, 3, padding=1, bias=True), nn.ReLU(inplace=True),
(3): nn.MaxPool2d(2, 2),  # [32, 32]
(4): nn.Conv2d(16, 32, 3, padding=1, bias=True), nn.ReLU(inplace=True),
(5): nn.MaxPool2d(2, 2),  # [16, 16]
(6): nn.Conv2d(32, 64, 3, padding=1, bias=True), nn.ReLU(inplace=True),
(7): nn.MaxPool2d(2, 2),  # [8, 8]
(8): nn.Conv2d(64, 128, 3, padding=1, bias=True), nn.ReLU(inplace=True),
(9): nn.MaxPool2d(2, 2),  # [4, 4]
(10): nn.Conv2d(128, 128, 1, padding=0, bias=True), nn.ReLU(inplace=True),
(11): nn.Linear(128 * 4 * 4, out_dim), nn.ReLU()
\end{lstlisting}

\vspace{0.5em}

We introduced a hyperparameter, denoted as BC loss threshold
, designed to terminate BC when a specified loss threshold is surpassed. This approach effectively prevents overfitting to a particular batch 
, allowing computational resources to be reallocated for sample collection and training on varied batches. This methodology significantly accelerates the training process in terms of wall time. Observations indicate that, following the initial updates, the utilization of the total BC iterations rarely reaches completion, often terminating after less than 10\% of the maximum allotted iterations. This is particularly evident during the early stages of training, where the agent, being in a nascent state of development, requires extensive training to attain any level of success.

For tasks in the ManiSkill , we use the following hyperparameters in Table \ref{tab:ManiSkill_S2V}:

\begin{table}[H]
    \centering
    \begin{tabular}{l@{\ \ \ \ }c}
         \toprule
         Name & Value \\
         \midrule
         Buffer size & 60k/300k\\
         Learning rate & 3e-4\\
         $N_{collect}$ & 64\\
         Batch size & 64/128\\
         Update-to-data ratio & 1\\
         BC loss threshold & 0.01\\
         \bottomrule
    \end{tabular}
    \caption{Hyperparameters for tasks of ManiSkill using \method}
     \label{tab:ManiSkill_S2V} 
\end{table}

For the OpenCabinetDrawer and MoveBucket tasks, which were more memory-intensive, we utilized a replay buffer of $60,000$ and a minibatch size of $64$. For the remaining tasks, we employed a larger replay buffer of 300,000 and increased the minibatch size to $128$.

For tasks in the DMControl, we use the following hyperparameters in Table \ref{tab:DMControl_S2V}:

\begin{table}[H]
    \centering
    \begin{tabular}{l@{\ \ \ \ }c}
         \toprule
         Name & Value \\
         \midrule
         Buffer size  & 500k\\
         Learning rate & 3e-4\\
         $N_{collect}$ & 2000\\
         Batch size & 100\\
         Update-to-data ratio & 1\\
         BC loss threshold & 0.025\\
         \bottomrule
    \end{tabular}
    \caption{Hyperparameters for tasks of DMControl using \method}
    \label{tab:DMControl_S2V} 
\end{table}

For tasks in the Adroit, we use the following hyperparameters in Table \ref{tab:Adroit_S2V}:

\begin{table}[H]
    \centering
    \begin{tabular}{l@{\ \ \ \ }c}
         \toprule
         Name & Value \\
         \midrule
         Buffer size & 500k\\
         Learning rate & 3e-4\\
         $N_{collect}$ & 64\\
         Batch size & 512\\
         Update-to-data ratio & 0.5\\
         BC loss threshold & 0.1\\
        \bottomrule
    \end{tabular}
    \caption{Hyperparameters for tasks of Adroit using \method}  
    \label{tab:Adroit_S2V} 
\end{table}

\subsection{Asymmetric Actor Critic}
For the tasks, OpenCabinetDrawer and MoveBucket in ManiSkill, the architecture of CNN that input channels equal to $12$ and output channels equal to $256$ is as:

\vspace{1em}

\begin{lstlisting}[basicstyle=\footnotesize\ttfamily, breaklines=true, frame=single, numbers=none]
(0): nn.Conv2d(in_channels, 16, 3, padding=1, bias=True), nn.ReLU(inplace=True),
(1): nn.MaxPool2d(2, 2),  # [25, 62]
(2): nn.Conv2d(16, 16, 3, padding=1, bias=True), nn.ReLU(inplace=True),
(3): nn.MaxPool2d(2, 2),  # [12, 31]
(4): nn.Conv2d(16, 32, 3, padding=1, bias=True), nn.ReLU(inplace=True),
(5): nn.MaxPool2d(2, 2),  # [6, 15]
(6): nn.Conv2d(32, 64, 3, padding=1, bias=True), nn.ReLU(inplace=True),
(7): nn.MaxPool2d(2, 2),  # [3, 7]
(8): nn.Conv2d(64, 128, 3, padding=1, bias=True), nn.ReLU(inplace=True),
(9): nn.Linear(128 * 3 * 7, out_dim)
\end{lstlisting}

\vspace{1em}

For other tasks in ManiSkill, it is as:

\vspace{1em}

\begin{lstlisting}[basicstyle=\footnotesize\ttfamily, breaklines=true, frame=single, numbers=none]
(0): nn.Conv2d(in_channels, 16, 3, stride=2, padding=1, bias=True), nn.ReLU(inplace=True), # [64, 64]
(1): nn.Conv2d(16, 16, 3, stride=2, padding=1, bias=True), nn.ReLU(inplace=True), # [32, 32]
(2): nn.Conv2d(16, 32, 3, stride=2, padding=1, bias=True), nn.ReLU(inplace=True), # [16, 16]
(3): nn.Conv2d(32, 64, 3, stride=2, padding=1, bias=True), nn.ReLU(inplace=True), # [8, 8]
(4): nn.Conv2d(64, 128, 3, stride=2, padding=1, bias=True), nn.ReLU(inplace=True), # [4, 4]
(5): nn.AdaptiveMaxPool2d((1, 1))
(6): nn.Linear(128, out_dim), nn.ReLU()
\end{lstlisting}

\vspace{1em}

For tasks in the DMControl, the architecture of CNN that input channels equal to $8$ and output channels equal to $256$ is as:

\vspace{1em}

\begin{lstlisting}[basicstyle=\footnotesize\ttfamily, breaklines=true, frame=single, numbers=none]
(0): nn.Conv2d(in_channel, 32, 3, stride=2), nn.ReLU(),
(1): nn.Conv2d(32, 32, 3, stride=1), nn.ReLU(),
(2): nn.Conv2d(32, 32, 3, stride=1), nn.ReLU(),
(3): nn.Conv2d(32, 32, 3, stride=1), nn.ReLU()
(4): flatten(1)
(5): nn.Linear(32 * 35 * 35, 256),
(6): nn.LayerNorm(256)
\end{lstlisting}

\vspace{1em}

For tasks in the Adroit, the architecture of CNN that input channels equal to $3$ and output channels equal to $256$ uses as:

\vspace{1em}

\begin{lstlisting}[basicstyle=\footnotesize\ttfamily, breaklines=true, frame=single, numbers=none]
nn.Conv2d(in_channels, 16, 3, padding=1, bias=True), nn.ReLU(inplace=True),
nn.MaxPool2d(2, 2),  # [64, 64]
nn.Conv2d(16, 16, 3, padding=1, bias=True), nn.ReLU(inplace=True),
nn.MaxPool2d(2, 2),  # [32, 32]
nn.Conv2d(16, 32, 3, padding=1, bias=True), nn.ReLU(inplace=True),
nn.MaxPool2d(2, 2),  # [16, 16]
nn.Conv2d(32, 64, 3, padding=1, bias=True), nn.ReLU(inplace=True),
nn.MaxPool2d(2, 2),  # [8, 8]
nn.Conv2d(64, 128, 3, padding=1, bias=True), nn.ReLU(inplace=True),
nn.MaxPool2d(2, 2),  # [4, 4]
nn.Conv2d(128, 128, 1, padding=0, bias=True), nn.ReLU(inplace=True),
nn.Linear(128 * 4 * 4, out_dim), nn.ReLU()
\end{lstlisting}

\vspace{1em}

For tasks in ManiSkill, we use the following hyperparameters in Table \ref{tab:ManiSkill_AAC}:

\begin{table}[H]
    \centering
    \begin{tabular}{l@{\ \ \ \ }c}
        \toprule
        Name & Value \\
        \midrule
        Buffer size & 200k/500k\\
        Discount factor & 0.8\\
        Batch size & 512\\
        Learning rate & 3e-4\\
        Update-to-data ratio & 0.25\\
        Autotune Entropy? & True\\
        \bottomrule
    \end{tabular}
    \caption{Hyperparameters for tasks of ManiSkill using Asymmetric Actor Critic}  
    \label{tab:ManiSkill_AAC} 
\end{table}

We used a $200$k replay buffer for OpenCabinetDrawer and MoveBucket (since they were more memory-consuming) and 500k for the other tasks.

For tasks in the DMControl, we use the following hyperparameters in Table \ref{tab:DMControl_AAC}:

\begin{table}[H]
    \centering
    \begin{tabular}{l@{\ \ \ \ }c}
        \toprule
        Name & Value \\
        \midrule
        Buffer size & 500k\\
        Discount factor & 0.99\\
        Batch size & 512\\
        Learning rate & 3e-4\\
        Update-to-data ratio & 0.25\\
        Autotune Entropy? & True\\
        \bottomrule
    \end{tabular}
    \caption{Hyperparameters for tasks of DMControl using Asymmetric Actor Critic}  
    \label{tab:DMControl_AAC} 
\end{table}

For tasks in the Adroit, we use the following hyperparameters in Table \ref{tab:Adroit_AAC}:

\begin{table}[H]
    \centering
    \begin{tabular}{l@{\ \ \ \ }c}
        \toprule
        Name & Value \\
        \midrule
        Buffer size & 500k\\
        Discount factor & 0.95\\
        Batch size & 512\\
        Learning rate & 3e-4\\
        Update-to-data ratio & 0.125\\
        Autotune Entropy? & True\\
        \bottomrule
    \end{tabular}
    \caption{Hyperparameters for tasks of Adroit using Asymmetric Actor Critic}  
    \label{tab:Adroit_AAC} 
\end{table}

\section{Experimental Supplementary}
\label{sec:appendix_drq}
\subsection{Comparison of Asymmetric Actor Critic and DrQ-v2}

\begin{figure*}[htbp!]
    \centering
    \includegraphics[width=\textwidth]{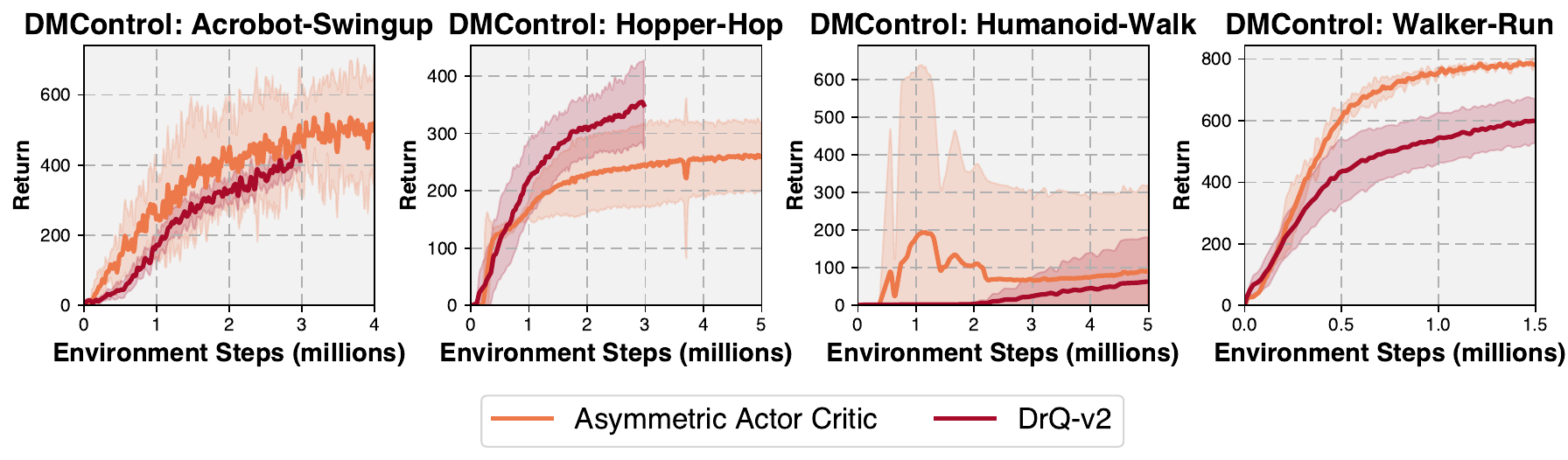}
    \caption{\textbf{Comparison of Asymmetric Actor Critic ~\cite{pinto2017asymmetric} and DrQ-v2 ~\cite{drq2}.} 
    We selected the Asymmetric Actor Critic as our primary visual RL method due to its capability to utilize low-dimensional state observations. 
    To validate the Asymmetric Actor Critic as a reasonable baseline, we compare it against DrQ-v2, which is a leading visual RL algorithm on DMControl. 
    Our comparison is conducted on four tasks that overlap between DrQ-v2's and our experiments, using learning curves directly downloaded from DrQ-v2's official repository.
    The environment steps used are consistent with Fig. \ref{fig:environment_steps}, utilizing the three seeds.
    The Asymmetric Actor Critic demonstrated competitive performance, surpassing DrQ-v2 in three of the four tasks, confirming its suitability as a representative for visual RL approaches.
    } 
    \label{fig:drq}
\end{figure*}

We compare the Asymmetric Actor Critic with DrQ-v2. The Asymmetric Actor Critic, selected for its effective use of low-dimensional state observations, is validated against DrQ-v2, a top performer on DMControl. Our evaluation covers four shared tasks using learning curves from DrQ-v2’s repository. As shown in Fig. \ref{fig:drq}, the Asymmetric Actor Critic outperforms DrQ-v2 in three of the four tasks, confirming its strength in visual RL.

\subsection{Details of Success Rate and Return with $95\%$ Confidence Interval}

\begin{table}[H]
    \centering
    \resizebox{\columnwidth}{!}{
    \begin{tabular}{l@{\ \ \ \ }cc@{\ \ \ \ }cc}
        \toprule
        & \multicolumn{2}{c}{\method} & \multicolumn{2}{c}{visual RL} \\
        \cmidrule(lr){2-3} \cmidrule(lr){4-5}
        Task Name & Mean & $95\%$ CI & Mean & $95\%$ CI \\
        \midrule
        ManisSkill: PickCube & 98.09 & [97.77, 98.41] & 99.64 & [99.07, 100.21] \\
        ManisSkill: StackCube & 95.82 & [95.26, 96.38] & 94.49 & [92.20, 96.78] \\
        ManisSkill: OpenDrawer & 95.66 & [93.45, 97.87] & 43.67 & [31.29, 56.05] \\
        ManisSkill: PickClutterYCB & 22.96 & [20.59, 25.33] & 6.27 & [4.86, 7.68] \\
        ManisSkill: TurnFaucet & 73.90 & [73.63, 74.17] & 71.62 & [56.97, 86.27] \\
        ManisSkill: PickSingleYCB & 59.69 & [51.78, 67.60] & 57.30 & [55.93, 58.67] \\
        ManisSkill: PegInsertion & 70.78 & [69.37, 72.19] & 0 & [0, 0] \\
        ManisSkill: MoveBucket & 75.33 & [73.17, 77.49] & 20.19 & [5.94, 34.44] \\
        DMControl: Walker-Run & 770.66 & [767.29, 774.03] & 812.00 & [811.11, 812.89] \\
        DMControl: Swimmer-6 & 508.71 & [487.61, 529.81] & 593.67 & [574.30, 613.04] \\
        DMControl: Hopper-Hop & 264.93 & [254.93, 274.93] & 260.59 & [218.92, 302.26] \\
        DMControl: Acrobot-Swingup & 536.64 & [520.04, 553.24] & 535.50 & [442.22, 628.78] \\
        DMControl: Humanoid-Walk & 296.29 & [269.78, 322.80] & 91.58 & [-63.37, 246.53] \\
        Adroit: Hammer & 99.03 & [98.38, 99.68] & 76.15 & [60.26, 92.04] \\
        Adroit: Pen & 83.66 & [81.53, 85.79] & 10.47 & [5.18, 15.76] \\
        Adroit: Relocate & 48.84 & [47.23, 50.45] & 0 & [0, 0] \\
        \bottomrule
    \end{tabular}
    }
    \caption{\textbf{Mean and $95\%$ Confidence Interval.} The table shows the mean return (DMControl), success rate (ManisSkill, Adroit), and $95\%$ confidence interval for each task across 3 seeds.}
    \label{tab:Adroit_AAC_3col_modified} 
\end{table}

\section{How We Tune Hyperparameters}
In this section, we detail the systematic approach taken to optimize hyperparameters in our experiments, as depicted in Fig. \ref{fig:environment_steps}, aiming to enhance sample efficiency and ensure fair comparisons across multiple benchmarks. Our tuning process was carried out on a benchmark-specific basis, focusing on representative tasks within each benchmark (e.g., StackCube, OpenDrawer, and PickClutterYCB from ManiSkill; all five tasks from DMControl; and Pen and Hammer from Adroit). The tuned hyperparameters were then consistently applied across other tasks within the same benchmark, ensuring both fairness and efficiency in our comparisons.

\subsubsection{Tuning Pipeline}
We began by identifying the most influential hyperparameters and tuned them sequentially, from the most to the least impactful. In cases where hyperparameters were highly interdependent, such as the imitation learning loss threshold and update-to-data ratio in \method, we tuned them together. This approach involved over 500 experiments, with some trials being manually stopped early when it was clear the hyperparameters were suboptimal.

\subsubsection{Key Findings}
\begin{itemize}
    \item \textbf{\method :} The imitation learning loss threshold and update-to-data ratio significantly influenced learning efficiency. We used manual coordinate descent to tune these, finding that optimal values varied across benchmarks.
    \item \textbf{$N_{collect}$:} This hyperparameter, indicating how frequently the policy network is trained, had a marginal impact on ManiSkill and Adroit but was crucial for DMControl due to its longer episode lengths.
    \item \textbf{Visual RL:} The discount factor was identified as the most critical parameter, with values of 0.8 for ManiSkill and 0.95 for Adroit yielding better results. Additionally, a smaller update-to-data ratio was preferred to avoid instability in RL training.
    \item \textbf{Shared Hyperparameters:} We used benchmark-specific CNN architectures, with slight adaptations for different resolutions and image numbers. The replay buffer size was maximized within memory constraints and adjusted for the varying visual observation sizes.
\end{itemize}

This tuning strategy not only enhanced performance but also offered insights into the key factors influencing different benchmarks. Further details will be made available with the code.

\section{Real-World Applicability for \method}

In this section, we discuss the real-world applicability of \method, which combines both visual and vector states. Although \method\ cannot be directly deployed in real-world scenarios, its practicality remains evident through the use of simulators. Previous research has demonstrated that simulators are effective tools for testing and refining algorithms, making them a critical component of \method\  \cite{loquercio2021learning, zhuang2023robot, lee2020learning}.

\subsection*{Training RL Agents in Simulators}
Training RL agents directly in the real world is often impractical due to high costs, safety risks, and operational challenges. Simulators provide a safer and more efficient alternative, enabling systematic comparisons between visual states (e.g., raw images) and vector states (e.g., object poses). These comparisons highlight the benefits of learning from structured representations.

\subsection*{Estimating Visual and Vector States in Real-World Scenarios}
In real-world applications, vector states can be derived from visual inputs using computer vision tools such as object detectors and pose estimators. These tools generate structured representations from perception modules, effectively bridging the gap between simulated and real-world environments.

\subsection*{Sim-to-Real Transfer}
Policies trained with \method\ can be deployed in real-world tasks through sim-to-real transfer techniques, such as domain randomization. This approach enhances robustness when transitioning from simulation to real-world conditions. The integration of visual inputs with structured representations further improves the reliability and interpretability of policies in practical robotic applications.

In summary, while our experiments are conducted in simulation, \method\ generalizes to real-world scenarios by leveraging perception tools, hybrid representations, and sim-to-real transfer methods.

\end{document}